\documentclass[conference]{IEEEtran}
\IEEEoverridecommandlockouts

\usepackage{tikz}
\usetikzlibrary{arrows.meta, positioning}
\usepackage{cite}
\usepackage{amsmath,amssymb,amsfonts}
\usepackage{algorithmic}
\usepackage{graphicx}
\usepackage{textcomp}
\usepackage{xcolor}
\usepackage{capt-of}
\usepackage{ragged2e}
\usepackage{float}     
\usepackage{placeins}  

\usepackage{graphics} 
\usepackage{epsfig} 
\usepackage{mathptmx} 
\usepackage{times} 
\usepackage{microtype}
\usepackage{algorithmic}
\usepackage{textcomp}
\usepackage{xcolor}
\usepackage{gensymb}
\usepackage{caption}
\usepackage{mwe}
\usepackage{verbatim}
\usepackage{amsbsy}
\usepackage{siunitx}
\usepackage{tabularx,booktabs}
\usepackage{dblfloatfix}
\usepackage{cite}
\usepackage{bm}
\usepackage{amsmath,amssymb,amsfonts}
\usepackage{subcaption}
\usepackage{float}

\usepackage{algorithm, algorithmic}
\usepackage[autolanguage]{numprint}

\usepackage{spreadtab}
\usepackage{multirow}
\usepackage{cite}
\usepackage{xcolor}
\usepackage{algorithmic}
\usepackage{textcomp}
\usepackage{xcolor}
\usepackage{gensymb}
\usepackage{caption}
\usepackage{mwe}
\usepackage{amsfonts}
\usepackage{amssymb}
\usepackage{verbatim}
\usepackage{amsbsy}
\usepackage{siunitx}
\usepackage{tabularx, booktabs}
\usepackage{soul}
\usepackage{tikz}
\usetikzlibrary{calc}

\usepackage[colorlinks=true,linkcolor=blue,citecolor=blue,urlcolor=blue]{hyperref}

\def\BibTeX{{\rm B\kern-.05em{\sc i\kern-.025em b}\kern-.08em
    T\kern-.1667em\lower.7ex\hbox{E}\kern-.125emX}}

\makeatletter
\newif\if@anonymize

\@anonymizefalse  

\if@anonymize
  \newcommand{\highlight@DoHighlight}{
    \fill [outer sep = -15pt, inner sep = 0pt, color=black]
          ($(begin highlight)+(0,8pt)$) rectangle ($(end highlight)+(0,-3pt)$) ;
  }

  \newcommand{\highlight@BeginHighlight}{
    \coordinate (begin highlight) at (0,0) ;
  }

  \newcommand{\highlight@EndHighlight}{
    \coordinate (end highlight) at (0,0) ;
  }

  \newdimen\highlight@previous
  \newdimen\highlight@current
  \newlength{\item@width}

  \DeclareRobustCommand*\anonymize{%
    \SOUL@setup
    \def\SOUL@preamble{%
      \begin{tikzpicture}[overlay, remember picture]
        \highlight@BeginHighlight
        \highlight@EndHighlight
      \end{tikzpicture}%
    }%
    \def\SOUL@postamble{%
      \begin{tikzpicture}[overlay, remember picture]
        \highlight@EndHighlight
        \highlight@DoHighlight
      \end{tikzpicture}%
    }%
    \def\SOUL@everyhyphen{%
      \discretionary{%
        \SOUL@setkern\SOUL@hyphkern
        \SOUL@sethyphenchar
        \tikz[overlay, remember picture] \highlight@EndHighlight ;%
      }{%
      }{%
        \SOUL@setkern\SOUL@charkern
      }%
    }%
    \def\SOUL@everyexhyphen##1{%
      \SOUL@setkern\SOUL@hyphkern
      \settowidth{\item@width}{##1}%
      \makebox[\item@width]{}%
      \discretionary{%
        \tikz[overlay, remember picture] \highlight@EndHighlight ;%
      }{%
      }{%
        \SOUL@setkern\SOUL@charkern
      }%
    }%
    \def\SOUL@everysyllable{%
      \begin{tikzpicture}[overlay, remember picture]
        \path let \p0 = (begin highlight), \p1 = (0,0) in \pgfextra
          \global\highlight@previous=\y0
          \global\highlight@current =\y1
        \endpgfextra (0,0) ;
        \ifdim\highlight@current < \highlight@previous
          \highlight@DoHighlight
          \highlight@BeginHighlight
        \fi
      \end{tikzpicture}%
      \settowidth{\item@width}{\the\SOUL@syllable}%
      \makebox[\item@width]{}%
      \tikz[overlay, remember picture] \highlight@EndHighlight ;%
    }%
    \SOUL@
  }
\else
  \newcommand{\anonymize}[1]{#1}
\fi

\newcommand{\linebreakand}{%
  \end{@IEEEauthorhalign}
  \hfill\mbox{}\par
  \mbox{}\hfill\begin{@IEEEauthorhalign}
}

\makeatother 

\begin{document}

\title{Mini Autonomous Car Driving based on 3D Convolutional Neural Networks}

\author{
    \anonymize{Pablo Moraes, Monica Rodriguez, Kristofer S. Kappel, Hiago Sodre,}\\ \anonymize{Santiago Fernandez, Igor Nunes, Bruna Guterres, Ricardo Grando} \\
    \IEEEauthorblockA{\anonymize{Technological University of Uruguay, UTEC, Uruguay}} \\
}

\maketitle

\begin{abstract}

Autonomous driving applications have become increasingly relevant in the automotive industry due to their potential to enhance vehicle safety, efficiency, and user experience, thereby meeting the increasing demand for sophisticated driving assistance features. However, the development of reliable and trustworthy autonomous systems has inherent challenges, such as high complexity, prolonged training periods, and intrinsic levels of uncertainty. Mini Autonomous Cars (MAC) are used as a practical testbed, allowing validation of autonomous control methodologies over small-scale setups. This simplified and cost-effective experimental setup facilitates rapid evaluation and comparison of machine learning models, which is particularly beneficial for algorithms requiring online training. To address these challenges and build upon this context of MACs, the present work offers a methodology based on RGB-D information and 3D Convolutional Neural Networks (3D CNNs) for MAC autonomous driving in simulated environments. We evaluate the performance of our proposed methodology with Recurrent Neural Networks (RNNs), whose architectures were trained and tested across two simulated tracks with distinct environmental features. Their performance was evaluated based on task completion success, lap-time metrics, and driving consistency. Results highlight how architectural modifications and track complexity influence the models’ generalization capability and vehicle control performance. The proposed version of 3D CNN has shown promising results when compared with RNNs. 

\vspace{1em}
\noindent\textbf{Video link:} \href{https://www.youtube.com/watch?v=6S2mfHWn3E8}{\texttt{https://youtu.be/6S2mfHWn3E8}}
\end{abstract}


\section{Introduction}

Autonomous vehicle technology has become a prominent research area within the automotive industry, driven by increasing market demand for integrating autonomous driving features into commercial vehicles \cite{prawira2025comparative}. This application field inherently requires solutions characterized by high levels of trustworthiness and reliability. Typically, available AI-based solutions for autonomous driving are complex, require extensive training periods and naturally involve certain levels of randomness or imprecision. Simulation environments have emerged as central tools in the evaluation of machine learning (ML)-based models. By employing a scaled-down version of the proposed problem, methodologies can be effectively presented and validated, facilitating the identification and assessment of relevant phenomena associated with the study. \cite{tiedemann2022miniature}. Therefore, Mini Autonomous Cars (MAC)\cite{moraes2024behavior, moraes2024urubots} may provide an experimental environment in which solutions validated in small-scale setups may be easily scaled up to meet more realistic conditions. MAC provides a flexible  and cost-effective setup for easily adapting experimental scenarios. By simplifying the creation and management of diverse conditions (e.g. lighting settings and obstacle configurations), it better supports extended evaluation of proposed autonomous driving methodologies and reduces the risks of damage.  

At the same time, the development of autonomous driving systems has increased consistently in recent decades thanks to advances in ML and Computer Vision \cite{elmquist2022software, samak2023autodrive}. Artificial Neural Networks (ANNs) have proven helpful for directly mapping images captured by front-facing cameras to vehicle control commands, enabling behavior cloning in controlled environments \cite{moraes2024behavior, search4}. However, these systems inherent challenges, such as high complexity, prolonged training periods, and intrinsic levels of uncertainty, which demands improved sensing for increase performance. Also, as training and validating models with minor changes in real-world settings may be complex, simulation has become a practical approach to developing and evaluating algorithms \cite{gymdonkeycar_repo, prawira2025comparative}. Therefore, complex sensing information, such as from RGB-D cameras, can increase the model's robustness, demanding more complex ANNs

\begin{figure}[H]
    \centering
    \includegraphics[width=\linewidth]{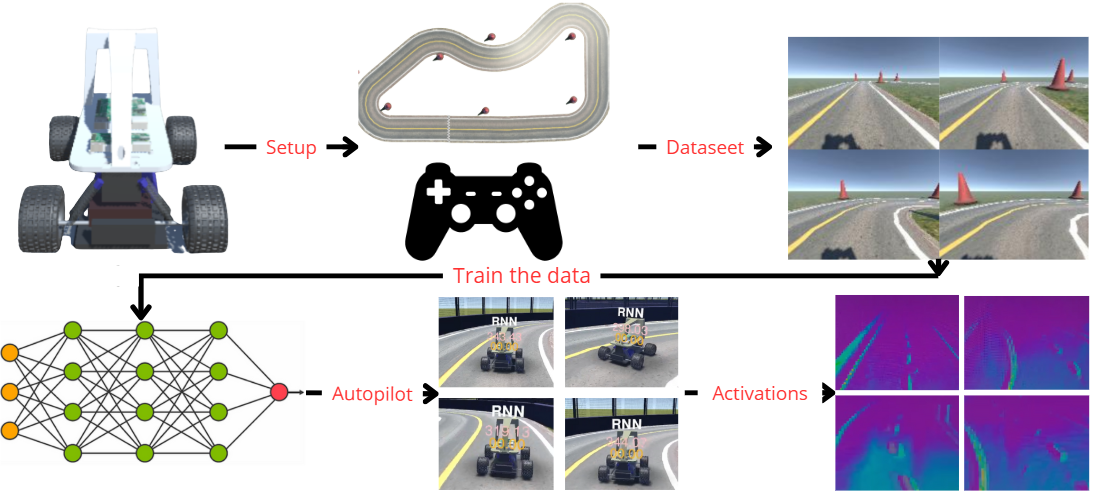}
    \caption{Block Diagram of the proposed methodology}
    \label{fig:system}
\end{figure}

Built upon this context, the present study provides a methodology for MAC autonomous driving based on 3D CNNs, performing a comparative assessment with RNNs within two simulated scenarios with distinct characteristics. Figure \ref{fig:system} presents the simplification of our proposed methodology. 

Specific datasets were generated for each track, and both standard and modified versions of each architecture were trained and tested. The performance was evaluated using objective metrics focused on time and consistency within the simulator. The results provide insights into how architectural changes and environmental complexity impact the networks’ ability to generalize and control autonomous vehicles, thereby facilitating the selection of effective models suited for future implementation on embedded hardware platforms. Overall, this paper presents the following contributions:

\begin{itemize}
    \item We show that our proposed method 3D CNN provides better performance in terms of average time to complete a proposed task when comparing with other RNN models.
 
    \item We provide an Ablation Study showing the impact of adding and removing layers in our proposed model, concluding that the version with the smallest amount of layers provided the best performance in terms of average time to complete a task.
\end{itemize}

The remainder of this paper is organized as follows: Section~\ref{sec:related} reviews related works addressing the proposed problem. Section~\ref{sec:methodology} describes the proposed methodology. Experimental results and discussions are reported in Section~\ref{sec:results}. Finally, Section~\ref{sec:conclusion} summarizes the concluding remarks and future directions.

\section{Related Works}
\label{sec:related}

Recent advances in autonomous driving research have increasingly focused on leveraging deep learning techniques, such as behavior cloning and reinforcement learning, to train control models in both simulated and real-world environments. Within this domain, a range of studies has investigated different model architectures, control strategies, and evaluation setups for improving the learning and generalization capabilities of autonomous systems.

Elmquist et al. (2022) \cite{elmquist2022software} describes a software framework, along with a hardware platform, used to design and analyze robotics autonomy algorithms in both simulation and real-world environments. Their physical platform is a 1/6-scale vehicle with reconfigurable components, allowing for testing of algorithms and sensor setups in both domains. Yildirim et al. (2024) \cite{academia21} validate perception systems for autonomous vehicles using behavior cloning for lateral control in real-world conditions, showing accurate steering angle predictions in real-time.

Codevilla et al. (2019) \cite{academia23} analyze the limitations of behavior cloning, especially in generalizing to dynamic objects and lacking a causal model, highlighting areas that need further research. Karnala and Campbell (2020) \cite{search11} investigate how model architecture affects vulnerability to adversarial attacks in machine learning-based autonomous driving systems.

Zhang and Du (2019) \cite{search6} propose a deep reinforcement learning method to train a scaled autonomous car in simulation, with transfer to the real world. Samak et al. (2023) \cite{samak2023autodrive} introduce AutoDRIVE, a modular ecosystem for research and education in autonomous driving that supports simulation and deployment of solutions including behavior cloning via deep imitation learning. Jung (2018) \cite{search4} uses behavior cloning to train a supervised CNN classifier on the DonkeyCar platform, optimizing autonomous driving in a realistic environment

Although previous studies have explored autonomous vehicle control using behavior cloning and deep learning models, such as RNNs and CNNs, there is a lack of detailed comparative evaluations between architectures, particularly between RNNs and 3D CNNs, in small-scale environments like Mini Autonomous Cars (MACs). Additionally, while the general performance and accuracy of models have been investigated, few studies analyze the impact of architectural adjustments (such as adding or removing layers) on model performance in visually complex environments. 

The present work addresses this gap by directly comparing RNN and 3D CNN models in the context of small autonomous vehicles (MACs) in a simulated environment. We also introduce an ablation study to examine how model complexity, modified by adding or removing layers, can improve performance in tracks with high visual complexity. The proposed methodology combines behavior cloning and simulation to develop autonomous driving systems based on prior works \cite{moraes2024behavior, elmquist2022software, samak2023autodrive, moraes2024urubots}. 

\section{Methodology}
\label{sec:methodology}

The proposed methodology based on 3D CNNs to investigate how architectural characteristics and environmental complexity impact the learning and control processes of autonomous vehicles, looking to apply these results to physical mini-autonomous vehicles in the future. It leverages open-source tools, such as MAC and DonkeyCar, as simulation testbeds for evaluating the performance of deep learning models under two distinct scenarios. 

The block diagram in Figure \ref{fig:system} illustrates the initial setup in which the simulated vehicle and environments were created and configured. Data from each simulation scenario were collected through manual driving, employing a joystick to control the virtual vehicle, thereby synchronizing front-camera images with corresponding control commands. This dataset provided the foundation for training our 3D CNNs models, including a comparison with standard and modified versions of Recurrent Neural Networks (RNNs). After the training phase, the models were deployed as autopilots to evaluate their performance in autonomous driving tasks and their generalization capabilities. The following subsections present detailed information regarding each stage of development and evaluation.

\subsection{Neural Network Architectures}

\begin{figure}[htbp]
    \centering
    \begin{minipage}[b]{0.50\linewidth}
        \centering
        \includegraphics[width=\linewidth]{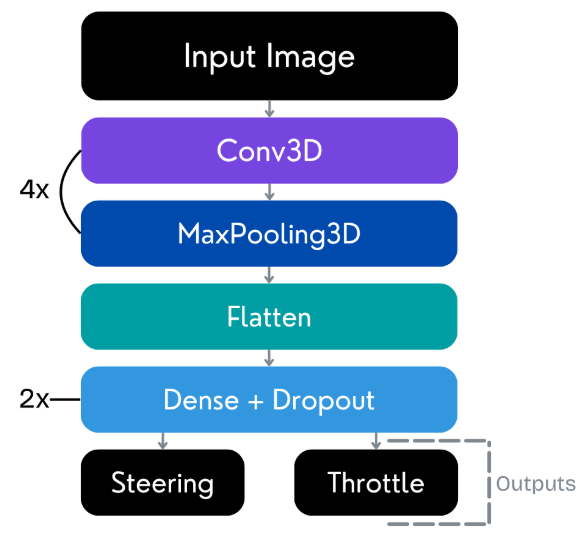}
        \caption{3D CNN from the framework}
        \label{fig:3dcnn_normal_placeholder}
    \end{minipage}%
    \hfill
    \begin{minipage}[b]{0.50\linewidth}
        \centering
       \includegraphics[width=\linewidth]{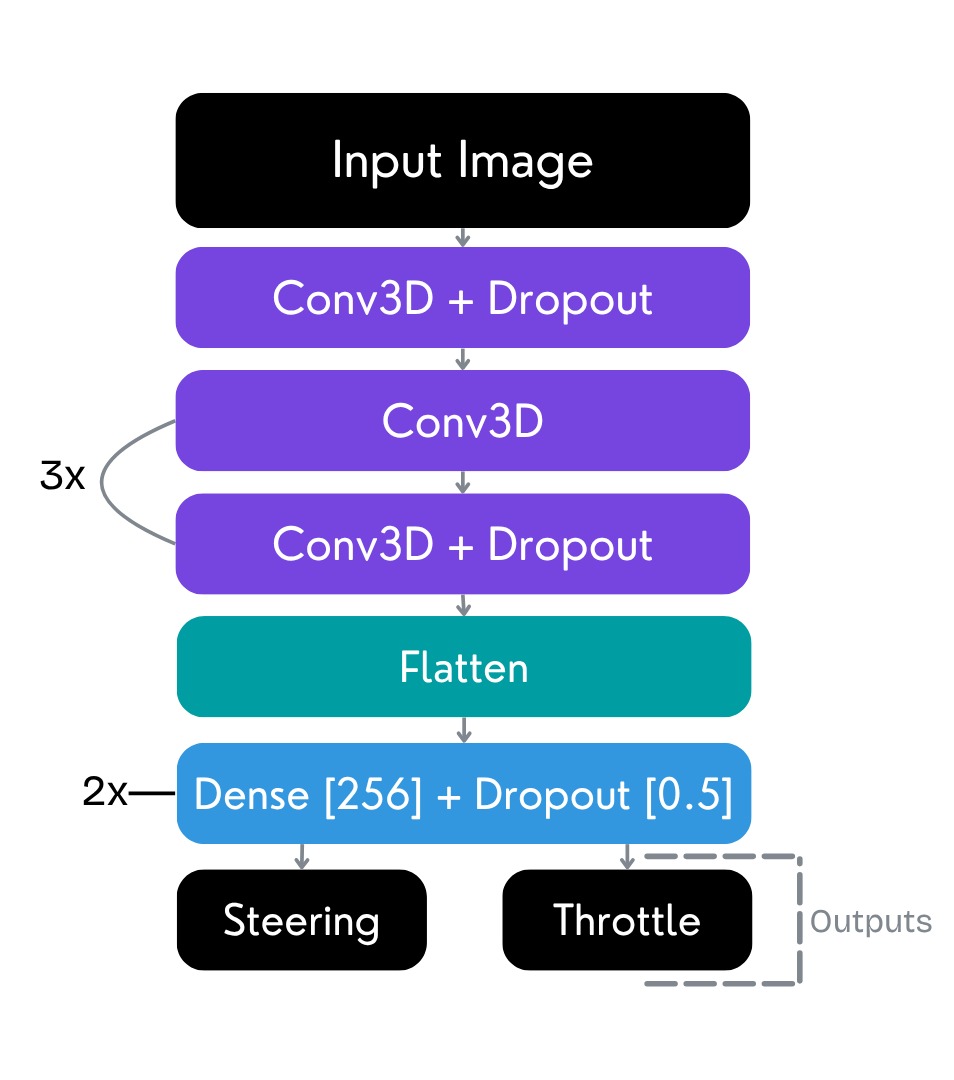}
        \caption{Modified 3DCNN.}
        \label{fig:3dcnn_modified_placeholder}
    \end{minipage}
\end{figure}

For this study, a main architecture based on 3D CNN were selected and compared with a RNN model. The Three-Dimensional Convolutional Neural Network (3D CNN), which extends convolutions into the temporal dimension, allowing joint spatial and temporal analysis of video sequences, shown in Figure \ref{fig:3dcnn_normal_placeholder}. This architecture facilitates detection of dynamic patterns across entire video sequences and has been successfully applied in autonomous driving simulations \cite{chu2020sim}.

\begin{figure}[htbp]
    \centering
    \begin{minipage}[b]{0.50\linewidth}
        \centering
        \includegraphics[width=\linewidth]{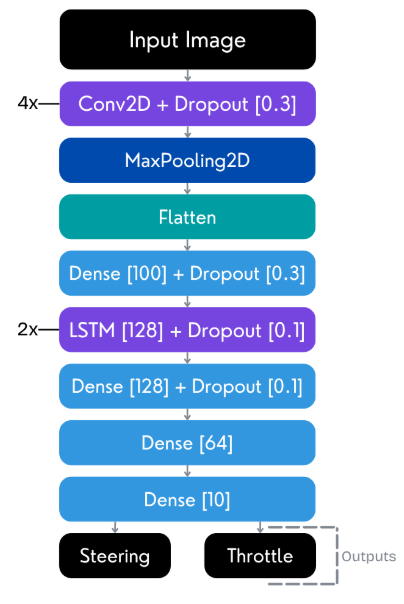}
        \caption{RNN-model from the framework\cite{donkeycar_repo}.}
        \label{fig:rnn_normal_placeholder}
    \end{minipage}%
    \hfill
    \begin{minipage}[b]{0.50\linewidth}
        \centering
        \includegraphics[width=\linewidth]{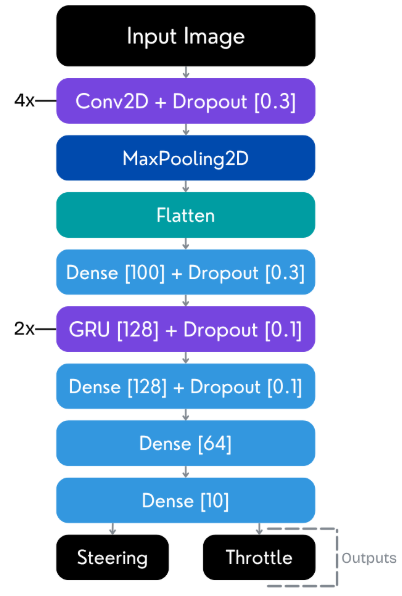}
        \caption{Modified RNN.}
        \label{fig:rnn_modified_placeholder}
    \end{minipage}
\end{figure}

The Recurrent Neural Network (RNN): designed to process temporal sequences of images and capture the dynamics of the environment, using Long Short-Term Memory (LSTM) layers in the standard version (see Figure \ref{fig:rnn_normal_placeholder}). These networks apply 2D convolutions to extract spatial features from each image, followed by LSTM layers that model long-term temporal dependencies through gating mechanisms, enabling the prediction of control commands based on the evolving visual environment \cite{kelchtermans2017hard}.

For each architecture, a standard and modified versions were trained, incorporating adjustments in architecture or parameters with the goal of improving performance. In particular, the modified RNN version replaces LSTM layers with Gated Recurrent Unit  (GRU) layers, which are known for their simpler structure and computational efficiency while maintaining comparable capacity for modeling temporal dependencies \cite{hochreiter1997long, cho2014learning, chung2014empirical} (see Figure \ref{fig:rnn_modified_placeholder}).

Mathematically, the GRU layers implement gating mechanisms that regulate the flow of information, defined by the following equations:
\begin{align}
z_t &= \sigma(W_z x_t + U_z h_{t-1}), \\
r_t &= \sigma(W_r x_t + U_r h_{t-1}), \\
\tilde{h}_t &= \tanh(W_h x_t + U_h (r_t \odot h_{t-1})), \\
h_t &= (1 - z_t) \odot h_{t-1} + z_t \odot \tilde{h}_t
\end{align}
where \( \sigma \) is the sigmoid function, \( \odot \) denotes element-wise multiplication, \( x_t \) is the input, and \( h_t \) the hidden state at time \( t \).

\begin{justify}
Regarding the 3D CNN networks, the modified version introduces residual blocks with LeakyReLU activation functions and Batch Normalization, following the principles of residual networks (ResNet) that ease training of deep models and improve generalization \cite{he2016deep, ioffe2015batch} (see Figure \ref{fig:3dcnn_modified_placeholder}).

Residual blocks implement a direct shortcut connection that sums the original input \( x \) to the output of the residual function \( F(x) \), expressed as:
\[
y = F(x, \{W_i\}) + x
\]
This structure helps mitigate degradation problems and facilitates gradient flow during backpropagation.

Additionally, Batch Normalization standardizes intermediate activations to stabilize and accelerate training, defined for an activation \( x \) in a mini-batch as:
\[
\hat{x} = \frac{x - \mu_B}{\sqrt{\sigma_B^2 + \epsilon}}, \quad
y = \gamma \hat{x} + \beta
\]
where \( \mu_B \) and \( \sigma_B^2 \) are the mini-batch mean and variance, and \( \gamma, \beta \) are learnable parameters.

These modifications aim to improve the learning capacity, robustness, and computational efficiency of the models to address the complexity of the simulated scenarios and facilitate future deployment on embedded hardware.
\end{justify}

\vspace{1em}

\vspace{1em}

\subsection{Simulation Scenarios}

Two simulated scenarios with distinct characteristics were selected to evaluate the performance of the autonomous driving models:

Mini Monaco: a closed circuit with walls on both sides, featuring sharp turns and layout variations that challenge the models' maneuvering and adaptation capabilities.
 
Generated Track: a shorter and simpler open track with cones randomly distributed outside the main route, designed to evaluate stability and speed under less demanding conditions.
 
\begin{figure*}[!h]
    \centering
    \begin{minipage}[t]{0.48\textwidth}
        \vspace{0pt}
        \includegraphics[width=\linewidth]{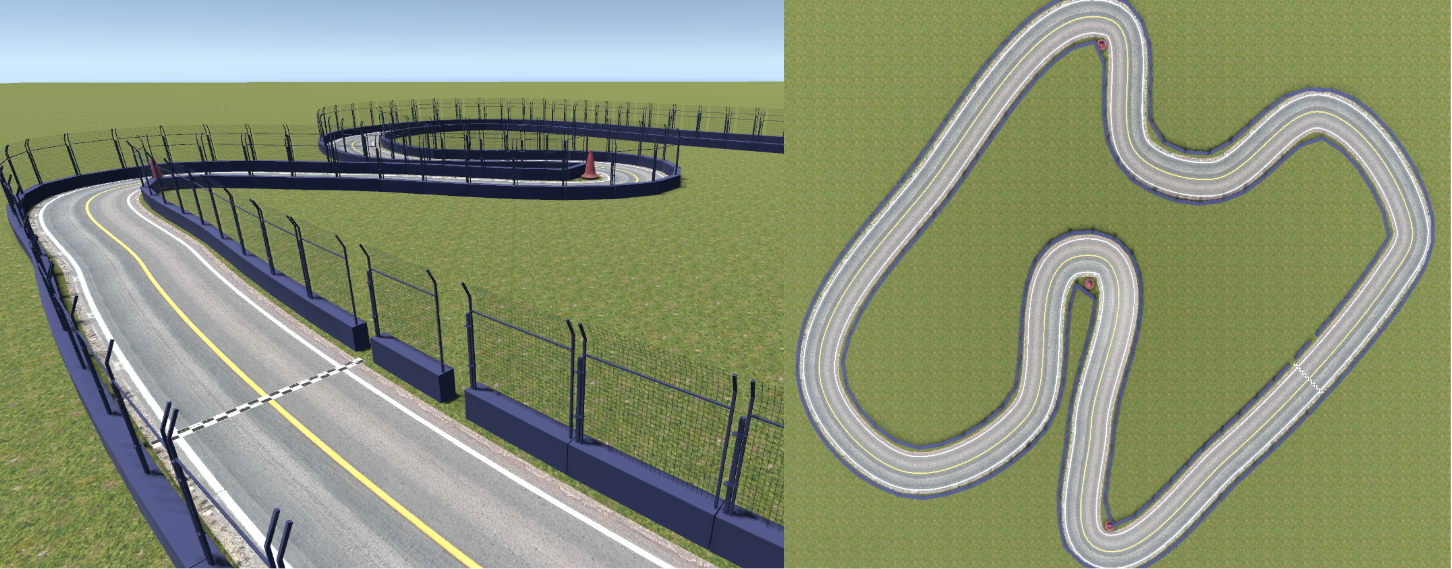}
        \caption{Combined view (aerial and track) of the Mini Monaco scenario.}
        \label{fig:monaco_completo}
    \end{minipage}%
    \hfill
    \begin{minipage}[t]{0.48\textwidth}
        \vspace{0pt}
        \includegraphics[width=\linewidth]{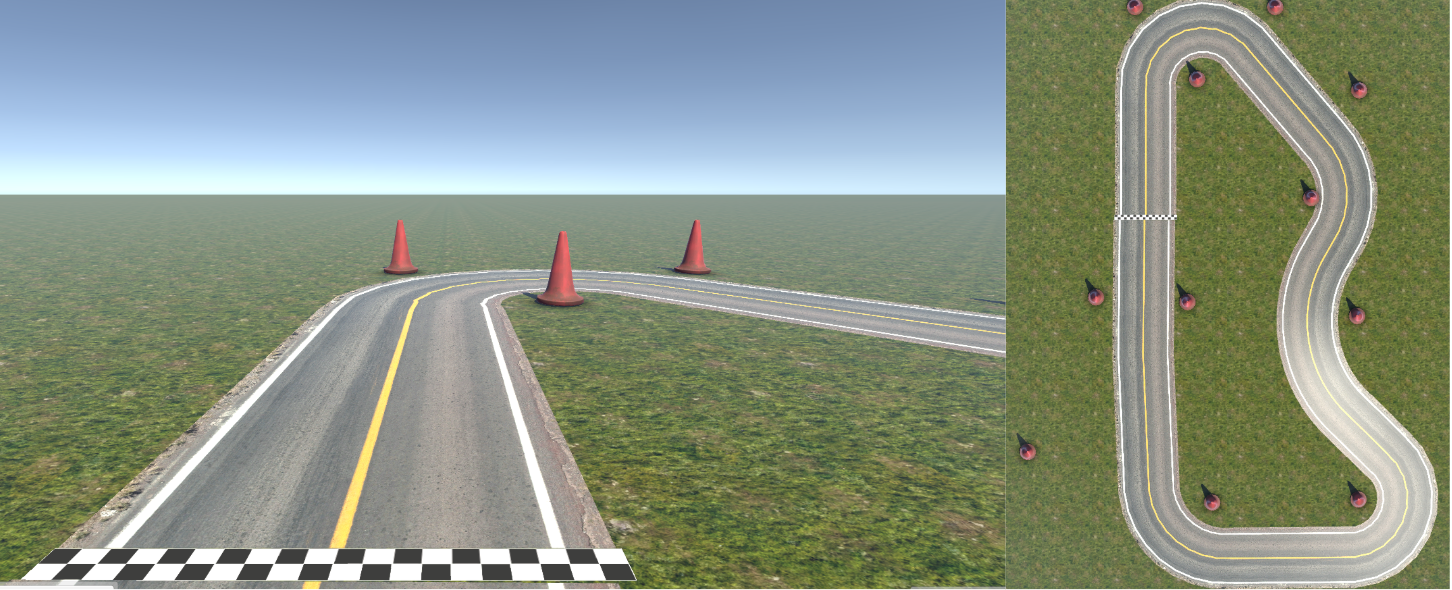}
        \caption{Combined view (aerial and track) of the Generated Track scenario.}
        \label{fig:generated_completo}
    \end{minipage}
\end{figure*}

Both scenarios were implemented in the DonkeyCar, a Unity-based simulated environment, leveraging open-source tools available on GitHub \cite{gymdonkeycar_repo, donkeycar_repo}. It offers visual and physical features similar to those of a real environment, facilitating the potential transfer of trained models to physical hardware. Figures \ref{fig:monaco_completo} and \ref{fig:generated_completo} shows both scenarios.

\subsection{Dataset Creation}

\begin{figure}[h]
    \centering
    \includegraphics[width=0.47\textwidth]{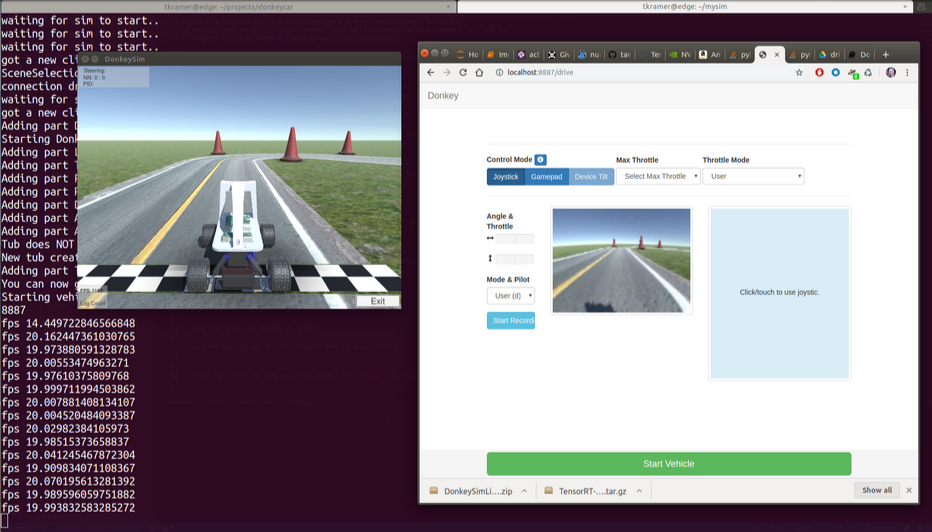}
    \caption{Donkey Sim configuration \cite{donkeycar_repo}}
    \label{fig:donkeyvista}
\end{figure}

The datasets for each scenario were generated by manual driving of the simulated standard vehicle, using a joystick connected to the DonkeyCar framework. The vehicle's front RGB-D camera images were captured synchronized with the control commands (steering angle and throttle).\ref{fig:donkeyvista} shows the simulation environment configuration.

\begin{minipage}[h]{0.47\textwidth}
    \centering
    \includegraphics[width=\linewidth]{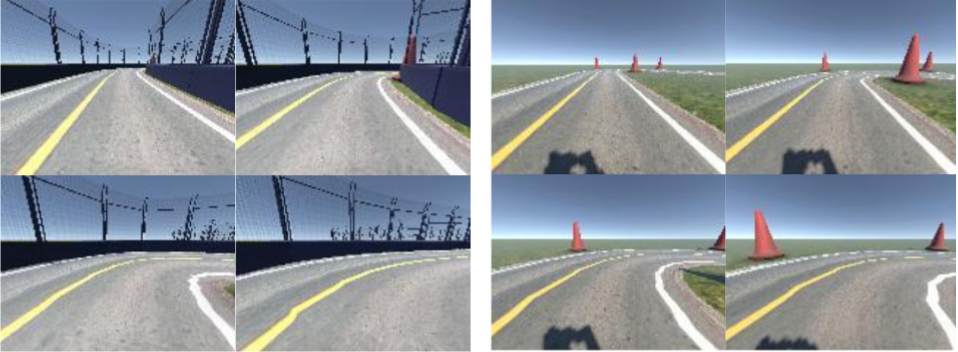}
    \captionof{figure}{View from the vehicle camera on both scenarios.}
    \label{fig:mini_timeframe}
\end{minipage}%

The camera system was configured to operate as a webcam capturing video at 20 frames per second (FPS), with a resolution of 160×120 pixels and three color channels (RGB) and a depth channel. It was synchronized with the control cycle
A total of 10,062 images were collected for Mini Monaco and 10,054 for Generated Track, forming representative datasets for each environment. For data splitting, 80\% was used for training and 20\% for validation, ensuring temporal and spatial consistency in each partition. The Figure \ref{fig:mini_timeframe} shows how the vehicle perceives each scenario, while Figures \ref{fig:dataset_mini_monaco} and \ref{fig:dataset_generated_track} present the control commands data (steering angle and throttle) collected by the user in the respective datasets.

\noindent
\begin{minipage}[t]{0.99\linewidth}
    \centering
    \includegraphics[width=\linewidth]{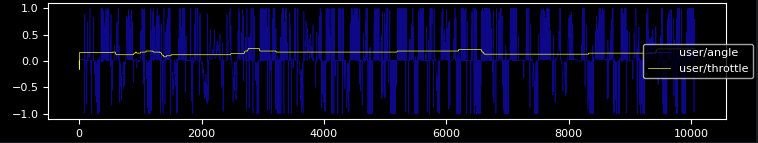}
    \captionof{figure}{Collected data: steering angle and throttle in Mini Monaco.}
    \label{fig:dataset_mini_monaco}
\end{minipage}

\vspace{1em} 

\noindent
\begin{minipage}[t]{0.99\linewidth}
    \centering
    \includegraphics[width=\linewidth]{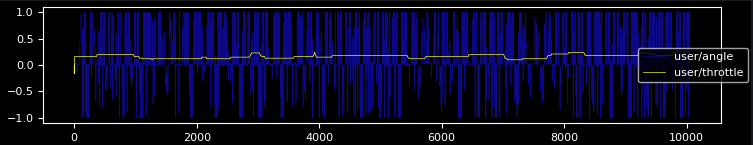}
    \captionof{figure}{Collected data: steering angle and throttle in Generated Track.}
    \label{fig:dataset_generated_track}
\end{minipage}

\subsection{Evaluation Metrics}

After training, the trained 3d CNN model was evaluated on two simulated tracks (Mini Monaco and Generated Track) and compared to the trained RNN model. Each model was tested over 30 laps per track, and the following metrics were used for evaluation:

\begin{itemize}
    \item Average Lap Time (sec): Time taken to complete each lap, averaged over 30 laps.
    \item Standard Deviation of Lap Times: Measure of the consistency in lap times.
    \item Track Deviations: Number of times the model went off-track or required major corrections.
    \item Response Speed (sec): Time is taken to adapt to track changes.
\end{itemize}

\section{Results}
\begin{figure}[h]
    \centering
    \includegraphics[width=0.9\linewidth]{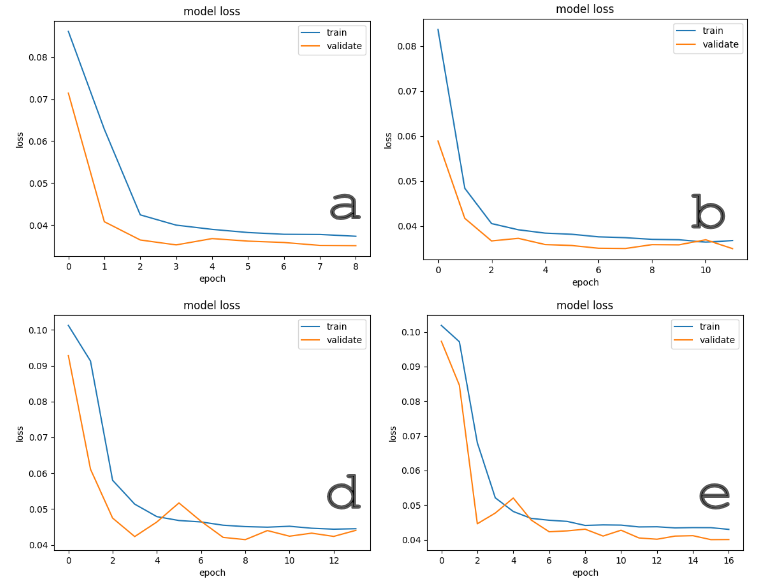}
    \caption{Training curves for RNN models: (a) RNN on Mini Monaco, (b) Modified RNN on Mini Monaco, (c) RNN on Generated Track, (d) Modified RNN on Generated Track.}
    \label{fig:juntos_rnn}
\end{figure}

\begin{justify}
\label{sec:results}

This section presents the evaluation of our trained models trained on the simulated tracks. The models were assessed in terms of stability (lap times), precision in track following, and response speed. Figures \ref{fig:juntos_rnn} and \ref{fig:juntos_3d} depict the training curves for the 3D CNN and for the RNN models, respectively, in their standard and modified versions, trained on both simulated tracks.

The following subsections present the performance of individual models for each track, along with a summary of the results and the outcomes of the ablation study.

\subsection{3D CNN Model (Figure \ref{fig:juntos_3d} (b))}

The 3D CNN Modified model struggled with unstable performance on Generated Track. Despite no major failures, the model exhibited clear signs of overfitting, with less ability to follow the track. Its lap times varied more than those of the RNN models.
The 3D CNN Modified model also underperformed on Mini Monaco. Although it maintained high speed, it struggled to stay on track during several laps, showing signs of overfitting to the training data. However, it adapted quickly during the initial laps.

\subsection{RNN model (Figure \ref{fig:juntos_rnn} (a))}

The RNN Default model followed the track smoothly with consistent lap times. It performed well in the curves and adapted effectively to layout variations. Track deviations were minimal, with no significant failures during the 30 laps.
On this track, the RNN Default model also displayed stable performance, maintaining consistent lap times throughout all 30 laps. The model followed the track with few errors and stayed on the path during every lap.
The Modified RNN, utilizing GRU layers, learned to navigate with slightly greater speed than the standard version. It showed improved consistency by staying close to the right side of the track, reducing the frequency of track departures compared to the RNN Default.
Similar to its performance on Mini Monaco, the Modified RNN demonstrated good speed and stability on Generated Track. The model showed consistency throughout all 30 laps, adapting well to the simpler layout, with no failures or significant variations.
\end{justify}

Overall, Table \ref{tab:tiempos_promedio} presents the average lap times for each model and track evaluated in this study. These values were obtained after 30 attempts per model on each simulated track.

\begin{table}[h]
    \centering
    \begin{tabular}{|c|c|c|}
    \hline
    \textbf{Model} & \textbf{Mini Monaco (sec)} & \textbf{Generated Track (sec)} \\
    \hline
    RNN Default & 36.24 (±0.56) & 14.75 (±0.22) \\
    RNN Modified & 36.56 (±0.48) & \textbf{14.45 (±0.18)} \\
    3D CNN Modified & \textbf{35.10 (±0.63)} & 14.58 (±0.25) \\
    \hline
    \end{tabular}
    \caption{Average times lap with standard deviation for the models evaluated on Mini Monaco and Generated Track.}
    \label{tab:tiempos_promedio}
\end{table}


\subsection{Ablation Study}

To further analyze our proposed method and the RNN version used to compare, we provide an Ablation Study. The first 3D CNN model version that was implemented did not achieve effective learning, as evidenced by its training curves, which indicated signs of overfitting. During testing on the Mini Monaco track—a scenario characterized by high visual complexity—the model performed poorly. It was unable to maintain a consistent trajectory and failed to respond adequately to visual cues in the environment. The vehicle ultimately collided with a wall, highlighting the model’s inability to anticipate and adapt to dynamic visual obstacles present on the track.

\begin{figure}[h]
    \centering
    \includegraphics[width=0.9\linewidth]{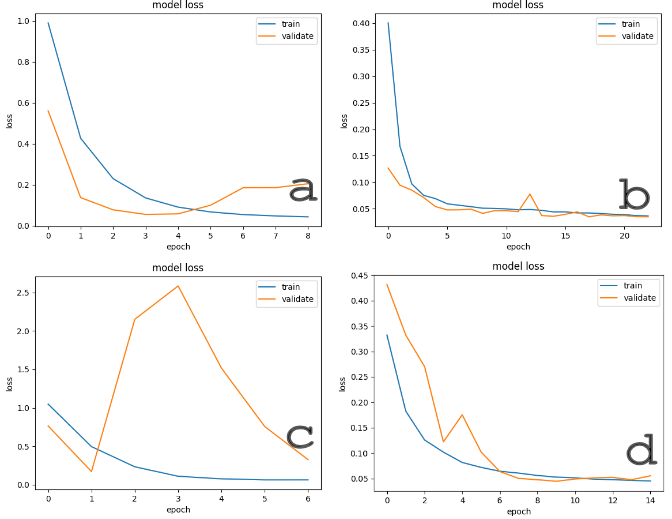}
    \caption{Training curves for 3D CNN models: (a) 3D CNN on Generated Track, (b) Modified 3D CNN on Generated Track, (c) 3D CNN on Mini Monaco, (d) Modified 3D CNN on Mini Monaco.}
    \label{fig:juntos_3d}
\end{figure}

To improve robustness, some modifications to the model were made, adjusting its architecture and removing some layers \ref{fig:3dcnn_normal_placeholder}, (\ref{fig:juntos_3d}, c). The modified model showed significant improvement in Mini Monaco, learning to follow the track with greater precision and stability \ref{fig:3dcnn_modified_placeholder}, (\ref{fig:juntos_3d}, d). However, despite this improvement, we decided to perform an Ablation Study to evaluate if adding or removing layers could further enhance its performance.

\begin{figure}[H]
\centering
\begin{minipage}[b]{0.5\linewidth}
    \centering
    \includegraphics[width=\linewidth]{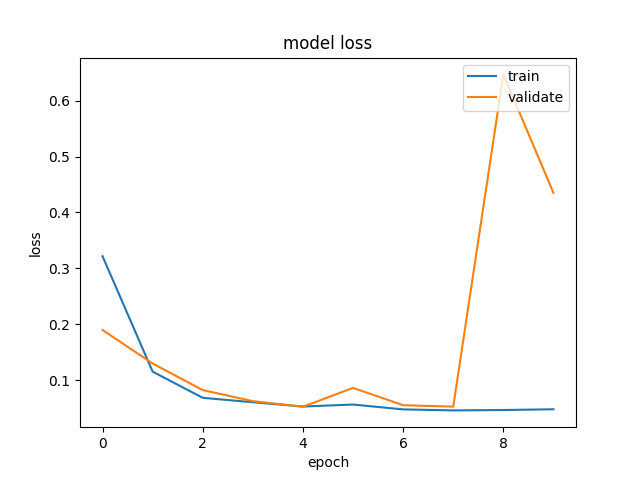}
    \caption{Model with one more 3D convolutional layer.}
    \label{fig:ablation_training_data_more_layer}
\end{minipage}%
\hfill
\begin{minipage}[b]{0.5\linewidth}
    \centering
    \includegraphics[width=\linewidth]{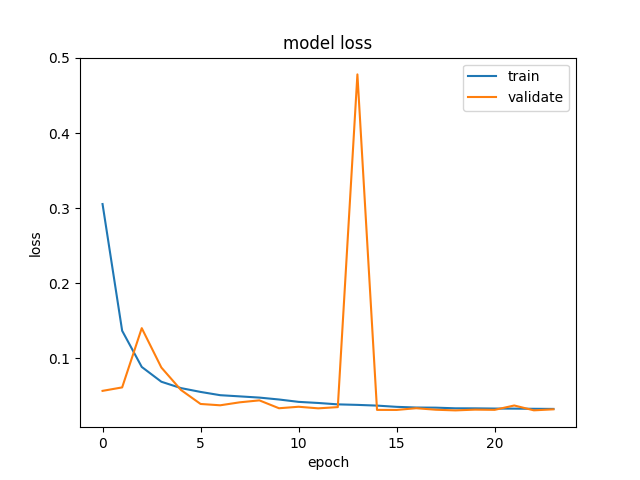}
    \caption{Model with one less 3D convolutional layer.}
    \label{fig:ablation_training_data_less_layer}
\end{minipage}
\end{figure}

The Ablation Study was conducted on the modified 3D CNN model. Two versions were tested: one with an additional 3D convolutional layer and another with one less 3D convolutional layer. The goal was to observe how these modifications affected the model’s ability to generalize and adapt to more complex scenarios, in this case, the Mini Monaco track. The training data can be seen in \ref{fig:ablation_training_data_less_layer}, \ref{fig:ablation_training_data_more_layer}.


Comparison of the performance of the three modified 3D CNN models can be seen in Table \ref{tab:desempeno_redes_modificadas_mini}, showing the average time lap on the Mini Monaco track for each of the trained models. Based on the results, we selected 3D CNN Modified version with one 3D convolutional layer less which was able to outperform the other structures.

\begin{table}[H]
    \centering
    \begin{tabular}{|c|c|}
    \hline
    \textbf{Model} & \textbf{Mini Monaco} \\
    \hline
    3D CNN Modified (1 layers less) & \textbf{34.10 (±0.63)} \\
    3D CNN Modified & 35.56 (±0.68) \\
    3D CNN Modified (1 layers more) & Overfit \\
    \hline
    \end{tabular}
    \caption{Modified 3DCNN on Mini Monaco.}
    \label{tab:desempeno_redes_modificadas_mini}
\end{table}

\section{Conclusions}
\label{sec:conclusion}
Based on the results obtained from the various models and scenarios, we can conclude that the 3D CNN models, particularly the Modified version, showed better performance on tracks with more complex visual features (Mini Monaco). These models, which integrate spatial and temporal feature extraction, learned the track’s dynamics effectively. However, overfitting was observed in simpler environments, such as the Generated Track, highlighting the need for fine-tuning the model’s complexity.

For the RNN Models, both the RNN Default and the Modified RNN (with GRU layers) demonstrated consistent performance across the simulated tracks. The RNN models showed good adaptability to changes in track layout and environmental factors, achieving stable lap times with minimal errors. However, their performance was limited by the lack of spatial-temporal feature extraction, which the 3D CNN models excel at.
    
For the last, our Ablation Study revealed that the model with one less layer (Modified 3D CNN) outperformed the others. It achieved faster lap times and better track-following precision, making it the most efficient model. The model with an additional layer exhibited overfitting, resulting in slower performance. This suggests that for the Mini Monaco track, a less complex model strikes the optimal balance between speed and accuracy.

This contribution provides a deeper understanding of how to design efficient machine-learning models for autonomous vehicles in complex environments, which may serve as a foundation for future research and applications in MACs.

Overall, the results indicate that reducing model complexity enhances performance in environments with numerous visual details. The best-performing model was the one with fewer layers, which showed faster speed and more consistent track-following, making it the most suitable choice for complex tracks like Mini Monaco. The RNN models, while more stable, did not outperform the 3D CNN models in more challenging tracks. Future research includes evaluating our real MAC and performing more evaluation with other ANNs and methods. 

\section*{Acknowledgements}

\anonymize{The authors would like to thank the Technological University
of Uruguay team and the UruBots robotics competitions team
in Rivera.}

\bibliographystyle{IEEEtran}
\bibliography{bibliography/IEEEexample}
\end{document}